\documentclass[10pt,twocolumn,letterpaper]{article}

\usepackage[pagenumbers]{cvpr} 

\usepackage{graphicx}
\usepackage{amsmath}
\usepackage{amssymb}
\usepackage{booktabs}

\usepackage{bbm}
\usepackage{multirow,array}
\usepackage{bm}
\usepackage{bbding}
\usepackage[accsupp]{axessibility}  

\usepackage[pagebackref,breaklinks,colorlinks]{hyperref}

\usepackage[capitalize]{cleveref}
\crefname{section}{Sec.}{Secs.}
\Crefname{section}{Section}{Sections}
\Crefname{table}{Table}{Tables}
\crefname{table}{Tab.}{Tabs.}

\def\cvprPaperID{4900} 
\def\confName{CVPR}
\def\confYear{2023}

\begin{document}

\title{Modeling Inter-Class and Intra-Class Constraints in Novel Class Discovery}

\author{Wenbin Li\textsuperscript{$1$},\quad Zhichen Fan\textsuperscript{$1$},\quad Jing Huo\textsuperscript{$1$}\thanks{Corresponding author},\quad Yang Gao\textsuperscript{$1$} \\
\textsuperscript{$1$}State Key Laboratory for Novel Software Technology, Nanjing University, China\quad
}

\maketitle

\begin{abstract}
Novel class discovery (NCD) aims at learning a model that transfers the common knowledge from a class-disjoint labelled dataset to another unlabelled dataset and discovers new classes (clusters) within it. Many methods, as well as elaborate training pipelines and appropriate objectives, have been proposed and considerably boosted performance on NCD tasks. Despite all this, we find that the existing methods do not sufficiently take advantage of the essence of the NCD setting. To this end, in this paper, we propose to model both inter-class and intra-class constraints in NCD based on the symmetric Kullback-Leibler divergence (sKLD). Specifically, we propose an inter-class sKLD constraint to effectively exploit the disjoint relationship between labelled and unlabelled classes, enforcing the separability for different classes in the embedding space. In addition, we present an intra-class sKLD constraint to explicitly constrain the intra-relationship between a sample and its augmentations and ensure the stability of the training process at the same time. We conduct extensive experiments on the popular CIFAR10, CIFAR100 and ImageNet benchmarks and successfully demonstrate that our method can establish a new state of the art and can achieve significant performance improvements, \eg, 3.5\%/3.7\% clustering accuracy improvements on CIFAR100-50 dataset split under the task-aware/-agnostic evaluation protocol, over previous state-of-the-art methods. Code is available at \url{https://github.com/FanZhichen/NCD-IIC}. 
\end{abstract}

\section{Introduction}
Deep learning has made great progress and achieved remarkable results in many computer vision fields, especially in image classification~\cite{background_AlexNet,background_VGGNet,background_GoogLeNet,background_ResNet,background_DenseNet}. Unfortunately, these successes of deep learning heavily rely on a large amount of fully labelled data for training. On the other hand, in many realistic scenarios, it is difficult to collect or to annotate such a large-scale dataset. To address this problem, a new paradigm of \textit{novel class discovery (NCD)} has been proposed and attracted increasing attention in recent years~\cite{DTC_ICCV2019,RS_ICLR2020,NCL_CVPR2021,UNO_ICCV2021,DualRank_NeurIPS2021,ComEx_CVPR2022}.

The goal of NCD is to train a classification model on a labelled dataset and simultaneously transfer the latent common knowledge to discover new classes (or clusters) in another unlabelled dataset. Different from \textit{semi-supervised learning}~\cite{background_MixMatch,background_ReMixMatch,background_FixMatch,background_SimMatch} that assumes the labelled and unlabelled datasets share the same label space, in the setting of NCD, the classes of the unlabelled dataset are disjoint with those of the labelled dataset, which is more challenging. In addition, NCD is also different from the generic \textit{clustering}~\cite{background_DEC,background_DeepCluster,background_IIC,background_SwAV} in that an additional labelled dataset is available in NCD. In general, for the standard clustering methods, the clustering results are not unique. That is to say, there may be multiple different and approximately correct results for a certain unlabelled dataset. In contrast, thanks to the available labelled dataset, NCD can eliminate the semantic ambiguity with the label guidance and finally makes the clustering be consistent with the real visual semantics~\cite{DTC_ICCV2019}. Clearly, NCD is more realistic and more practical than unsupervised clustering.

In general, the existing NCD methods can be roughly divided into two categories, \ie, two-stage based methods and single-stage based methods~\cite{SpacingLoss_CVPRWorkshop2022}. Most of the early methods are two-stage by using labelled and unlabelled data in different stages, such as KCL~\cite{KCL_ICLR2018}, MCL~\cite{MCL_ICLR2019} and DTC~\cite{DTC_ICCV2019}. Typically, the two-stage based methods first learn an embedding network on the labelled set through supervised learning and then use it on the unlabelled set to discover new clusters with little modifications. In contrast, the latest methods are almost single-stage, such as, RS~\cite{RS_ICLR2020}, NCL~\cite{NCL_CVPR2021}, UNO~\cite{UNO_ICCV2021}, DualRank~\cite{DualRank_NeurIPS2021} and ComEx~\cite{ComEx_CVPR2022}, which use both labelled and unlabelled data in a single stage at the same time. The single-stage based methods can learn the feature representation and discover novel classes simultaneously, iteratively updating the learned feature embedding network and clustering results during the training process. Compared with the two-stage based methods, the single-stage methods can make more effective use of the similarity between labelled and unlabelled classes to achieve a better knowledge transfer between these two datasets. Therefore, in this paper, we will mainly focus on the single-stage direction of NCD.

However, we find that the current single-stage based NCD methods do not sufficiently take advantage of the essence of the NCD setting, that is to say, overlooking the disjoint characteristic between the labelled and unlabelled classes. In this sense, on one hand, the labelled and unlabelled samples cannot be effectively separated, weakening the discriminability of the learned features. On the other hand, because the labelled data is learned under supervision while the unlabelled data has no supervision, \ie, an imbalanced learning process (learning with different supervision strengths), it will make the learned feature representations biased toward the labelled data. In addition, we notice that although some methods~\cite{RS_ICLR2020,DualRank_NeurIPS2021} have used data augmentation to generate additional samples and gained significant performance improvements, they generally employ the mean squared error (MSE) as the consistency regularization, which cannot constrain the consistency well with a good generalization ability.

To address the above two issues, we propose to model both \textit{Inter-class and Intra-class Constraints} (IIC for short) built on the \textit{symmetric Kullback-Leibler divergence (sKLD)} for discovering the novel classes. To be specific, an \textit{inter-class sKLD constraint} is proposed to explicitly learn to separate different classes between labelled and unlabelled data, enhancing the discriminability of learned feature representations. Moreover, an \textit{intra-class sKLD constraint} is presented to fully learn the intra-relationship between samples and their augmentations. According to our experiments, such an intra-class sKLD constraint can also stable the training process in the training phase. We have conducted extensive experiments on three benchmarks, including CIFAR10~\cite{CIFAR}, CIFAR100~\cite{CIFAR} and ImageNet~\cite{ImageNet}, and show that the proposed two constraints can significantly and consistently outperform the existing novel class discovery methods by a large margin.

To summarize, our contributions are as follows:
\begin{itemize}
\item We propose a new \textit{inter-class Kullback-Leibler divergence constraint} to sufficiently model the relationship between the labelled and unlabelled datasets to learn more discriminative feature representations, which is somewhat overlooked in the literature. 
\item We propose a new \textit{intra-class Kullback-Leibler divergence constraint} to effectively exploit the relationship between a sample and its different transformations to learn invariant feature representations.
\item We evaluate the proposed constraints on three benchmark datasets for novel class discovery and obtain significant performance improvements over the state-of-the-art methods, which successfully demonstrates the effectiveness of the proposed method.
\end{itemize}

\section{Related Work}
Novel class discovery (NCD) is a new task attracted wide attention in recent years, which aims at discovering new classes in an unlabelled dataset given a class-disjoint labelled dataset as supervision. A variety of advanced NCD methods have been proposed and have tangibly improved the clustering performance on multiple benchmark datasets.

The early methods of NCD include KCL~\cite{KCL_ICLR2018}, MCL~\cite{MCL_ICLR2019} and DTC~\cite{DTC_ICCV2019}. In general, these methods first learn an embedding network of feature representations on the labelled data, and then use it directly for the unlabelled data. Specifically, KCL and MCL propose a framework for both cross-domain and cross-task transfer learning that leverages the pairwise similarity to represent categorical information, and learn the clustering network based on the pairwise similarity prediction through different objective functions, respectively. DTC extends the deep embedding clustering method~\cite{background_DEC} into a transfer learning setting and proposes a two-stage method. Importantly, Han \etal~\cite{DTC_ICCV2019} formalize the task of \textit{novel class discovery} for the first time.

Since then, the current NCD methods~\cite{RS_ICLR2020,NCL_CVPR2021,UNO_ICCV2021,Joint_ICCV2021,OpenMix_CVPR2021,DualRank_NeurIPS2021,SpacingLoss_CVPRWorkshop2022,ComEx_CVPR2022} are almost single-stage and can take greater advantage of both labelled and unlabelled data. RS~\cite{RS_ICLR2020} introduces a three-step learning pipeline, which first trains the representation network with all labelled and unlabelled samples using self-supervised learning, and then uses ranking statistics to obtain pairwise similarity between unlabelled samples, and finally use the pairwise similarity to discover novel classes. DualRank~\cite{DualRank_NeurIPS2021} expands RS to a two-branch framework from both global and local levels. Similarly, DualRank uses dual ranking statistics and mutual knowledge distillation to generate pseudo labels and ensure the consistency between two branches. In order to generate pairwise pseudo labels, Joint~\cite{Joint_ICCV2021} employs a Winner-Take-All (WTA) hashing algorithm~\cite{background_WTA} on the shared feature space for NCD.

NCL~\cite{NCL_CVPR2021} and OpenMix~\cite{OpenMix_CVPR2021} are largely motivated by contrastive learning~\cite{background_SimCLR, background_MoCo} and Mixup~\cite{background_Mixup}, respectively. NCL introduces the contrastive loss to learn more discriminative representations. On the other hand, OpenMix uses Mixup to mix labelled and unlabelled samples, building a learnable relationship between the two parts of data. Instead of using multiple objectives, UNO~\cite{UNO_ICCV2021} introduces a unified objective function to transfer knowledge from the labelled set to unlabelled set. More recently, Joseph \etal~\cite{SpacingLoss_CVPRWorkshop2022} categorize the existing NCD methods into two classes (\ie, two- and single-stage based methods), according to whether the labelled and unlabelled samples are available at the same time or not. They also propose a spacing loss to enforce separability between labelled and unlabelled points in the embedding space. ComEx~\cite{ComEx_CVPR2022} focuses on the generalized NCD (GNCD), aka generalized category discovery (GCD)~\cite{GCD_CVPR2022}, and proposes two groups of compositional experts to solve this problem.

To our best knowledge, the existing approaches do not make full use of the disjoint characteristic between labelled and unlabelled classes. In addition, we find that some methods utilizes the mean squared error (MSE) to constrain the learned representations of data augmentation, but it cannot achieve the desired effect. Instead, we propose inter-class and intra-class constraints based on the symmetric Kullback-Leibler divergence (sKLD) for NCD.

\section{Method}
\subsection{Problem Formulation}
In the NCD setting, given a labelled dataset $\mathcal{D}^l=\{(\bm{x}^l_1,y^l_1),\ldots,(\bm{x}^l_N,y^l_N)\}$, the goal is to automatically discover $C^u$ clusters (or classes) in an unlabelled dataset $\mathcal{D}^u=\{\bm{x}^u_1,\ldots,\bm{x}^u_M\}$, where each $\bm{x}^l_i$ in $\mathcal{D}^l$ or $\bm{x}^u_i$ in $\mathcal{D}^u$ is an image and $y^l_i\in\mathcal{Y}=\{1,\ldots,C^l\}$ is the corresponding class label of $\bm{x}^l_i$. In particular, we assume that the set of $C^l$ labelled classes is disjoint with the set of $C^u$ unlabelled classes. In this sense, the core of NCD is how to effectively learn transferable semantic knowledge from the disjoint labelled dataset $\mathcal{D}^l$ to help performing clustering on the unlabelled dataset $\mathcal{D}^u$. Following the literature~\cite{RS_ICLR2020, NCL_CVPR2021, UNO_ICCV2021}, we also assume the number of unlabelled classes $C^u$ is known a priori in this paper. 

To tackle this challenge problem, we propose two \textit{symmetric Kullback-Leibler divergence (sKLD)} based constraints from both \textit{inter-class} and \textit{intra-class} perspectives to learn more discriminative feature representations for NCD models (see~\cref{framework}). In the following sections, we first introduce the \textit{inter-class sKLD constraint} and \textit{intra-class sKLD constraint} in~\cref{sec:skld}, and then summarize the overall objective for training in~\cref{sec:overall_objective}.

\subsection{Symmetric Kullback-Leibler Divergence for Novel Class Discovery}
\label{sec:skld}
According to the above analyses, to effectively utilize the two parts of data in NCD, \ie, a labelled set and an unlabelled set, we develop two \textit{inter-class and intra-class symmetric Kullback-Leibler divergence (sKLD) constraints} to better accomplish the NCD task, especially in the single-stage paradigm. 

\begin{figure}[!t]
\centering
\includegraphics[width=1\columnwidth]{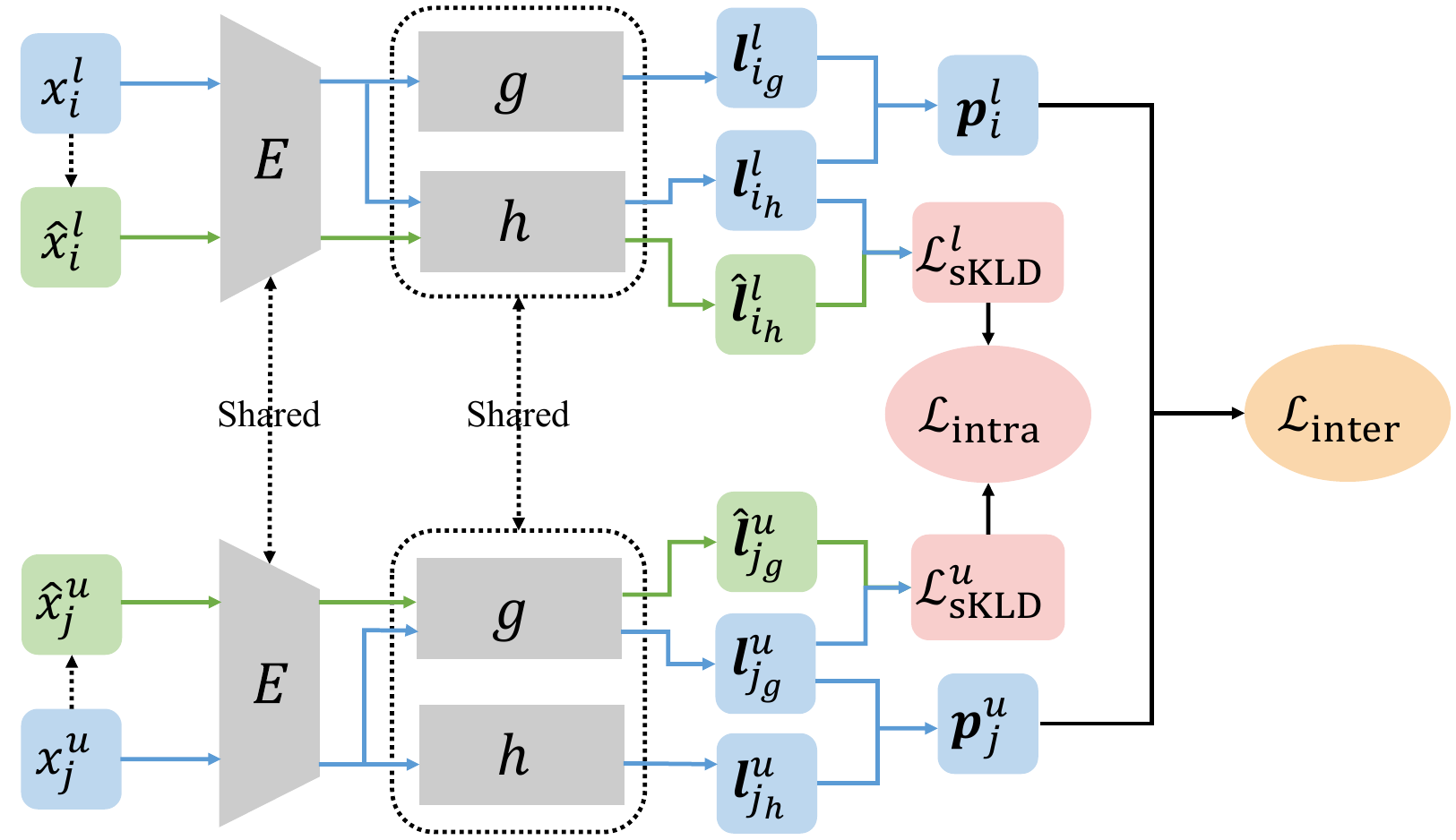}
\caption{Architecture of the proposed method. We present the ``raw samples'' (labelled sample $\bm{x}^l_i$, unlabelled sample $\bm{x}^u_j$, corresponding logits and probability distributions) in blue and the ``augmented counterparts'' (labelled counterpart $\hat{\bm{x}}^l_i$, unlabelled counterpart $\hat{\bm{x}}^u_j$, corresponding logits and probability distributions) in green. The samples and their augmentations are inputted into the shared encoder $E$, and then fed into two classification heads $h$ and $g$, obtaining predictions to calculate both inter-class sKLD $\mathcal{L}_\text{inter}$ and intra-class sKLD $\mathcal{L}_\text{intra}$. For brevity, we omit the calculation process of the standard cross-entropy loss $\mathcal{L}_\text{CE}$.}
\label{framework}
\end{figure}

Following UNO~\cite{UNO_ICCV2021}, as shown in~\cref{framework}, the architecture of our model consists of two parts: an encoder $E$ and two classification heads $h$ and $g$. The encoder $E$ is implemented as a standard convolutional neural network (CNN), which converts an input images into a feature vector. The head $h$ belonging to the labelled data is implemented as a linear classifier with $C^l$ output neurons, and the head $g$ belonging to the unlabelled data is composed of a multi-layer perceptron (MLP) and a linear classifier with $C^u$ output neurons. In the training phase, each sample $\bm{x}_i$ will be first encoded as a feature vector by $E$, and then will be passed through both classification heads to obtain the corresponding logits $\bm{l}_{i_h}\in\mathbb{R}^{C^l}$ and $\bm{l}_{i_g}\in\mathbb{R}^{C^u}$, respectively. After that, the two logits are concatenated together as $\bm{l}_i=[\bm{l_{i_h}},\bm{l_{i_g}}]\in\mathbb{R}^{C^l+C^u}$ and fed into a softmax layer $\sigma$ with a temperature $\tau$, obtaining the probability distribution $\bm{p}_i=\sigma(\bm{l}_i/\tau)$.

\textbf{Inter-class Symmetric KLD Constraint.}
When solving the NCD problem, in the pipeline of the single-stage based methods, both labelled and unlabelled images will be accessed in each mini-batch during training. Although the distributions of these two parts of images are similar, in fact, the representations of them should be different from each other as much as possible, \ie, the separability between the labelled and unlabelled classes. However, this point is somewhat overlooked in the existing single-stage based methods. Therefore, to address this issue, we propose an \textit{inter-class sKLD constraint} to explicitly enlarge the distance between each labelled sample and each unlabelled sample in the current mini-batch using a symmetric Kullback-Leibler divergence distance. The formulation between a pair of samples is as follows:
\begin{equation}
\label{inter_1}
    \mathcal{L}_\text{sKLD}=\frac{1}{2}\big(D_\text{KL}(\bm{p}^l_i||\bm{p}^u_j)+D_\text{KL}(\bm{p}^u_j||\bm{p}^l_i)\big),
\end{equation}
where $\bm{p}^l_i$ and $\bm{p}^u_j$ are the probability distributions generated for the labelled image $\bm{x}^l_i|_{i=1}^N$ and unlabelled image $\bm{x}^u_j|_{j=1}^M$ in a mini-batch, respectively. $D_\text{KL}$ is the Kullback-Leibler (KL) divergence defined as
\begin{equation}
\label{inter_2}
    D_{\rm KL}(\bm{p}^l_i||\bm{p}^u_j)=\sum^{C^l+C^u}_{k=1}\bm{p}^l_i(k)\log\frac{\bm{p}^l_i(k)}{\bm{p}^u_j(k)},
\end{equation}
\begin{equation}
\label{inter_3}
    D_{\rm KL}(\bm{p}^u_j||\bm{p}^l_i)=\sum^{C^l+C^u}_{k=1}\bm{p}^u_j(k)\log\frac{\bm{p}^u_j(k)}{\bm{p}^l_i(k)}.
\end{equation}
Therefore, the inter-class sKLD over a mini-batch is
\begin{equation}
\label{inter_4}
\mathcal{L}_\text{inter-class}=\frac{1}{NM}\sum^N_{i=1}\sum^M_{j=1}\mathcal{L}_{\rm sKLD},
\end{equation}
where $N$ and $M$ are the numbers of labelled and unlabelled images in a mini-batch, respectively.

\textbf{Intra-class Symmetric KLD Constraint.}
Recently, some NCD methods~\cite{UNO_ICCV2021,OpenMix_CVPR2021} have introduced data-augmentation techniques, normally used in self-supervised learning~\cite{background_RotNet}, to help improve the clustering accuracy when discovering novel classes. In particular, UNO~\cite{UNO_ICCV2021} leverages a multi-view strategy built on data augmentation and uses a swapped prediction mechanism through the Sinkhorn-Knopp algorithm~\cite{SK} to generate pseudo labels for augmented unlabelled images. However, UNO does not consider the intra-class constraint between the augmented images, no matter labelled nor unlabelled. We argue that for different views (augmentations) of the same image, the outputs produced by the corresponding classification heads should be consistent. That is to say, the distance between any two probability distributions of different augmentations of the same image should be small.

Note that, because the logits of unlabelled images passing through the labelled head $h$ and the logits of labelled images passing through the unlabelled head $g$ will gradually approach to a zero vector, it is not necessary to calculate the consistency regularization loss for these two kinds of logits. In this paper, we only consider the logits of labelled images generated by the labelled head $h$ and the logits of unlabelled images generated by the unlabelled head $g$:
\begin{equation}
    \mathcal{L}_\text{sKLD}^l=\frac{1}{2}\big(D_\text{KL}(\bm{p}^l_{i_h}||\hat{\bm{p}}^l_{i_h})+D_\text{KL}(\hat{\bm{p}}^l_{i_h}||\bm{p}^l_{i_h})\big),
\end{equation}
\begin{equation}
    \mathcal{L}_\text{sKLD}^u=\frac{1}{2}\big(D_\text{KL}(\bm{p}^u_{j_g}||\hat{\bm{p}}^u_{j_g})+D_\text{KL}(\hat{\bm{p}}^u_{j_g}||\bm{p}^u_{j_g})\big),
\end{equation}
\begin{equation}
    \mathcal{L}_\text{intra-class}=\frac{1}{N}\sum^N_{i=1}\mathcal{L}^l_\text{sKLD}+\frac{1}{M}\sum^M_{j=1}\mathcal{L}^u_\text{sKLD},
\end{equation}
where $\bm{p}^l_{i_h}\in\mathbb{R}^{C^l}$ and $\hat{\bm{p}}^l_{i_h}\in\mathbb{R}^{C^l}$ indicate the probability distributions of the labelled image $\bm{x}^l_i$ and its augmentation $\hat{\bm{x}}^l_i$ through the labelled head $h$ and the softmax layer $\sigma$, respectively. Similarly, $\bm{p}^u_{j_g}\in\mathbb{R}^{C^u}$ and $\hat{\bm{p}}^u_{j_g}\in\mathbb{R}^{C^u}$ represent the probability distributions corresponding to the unlabelled image $\bm{x}^u_j$ and its augmentation $\hat{\bm{x}}^u_j$, respectively.

\subsection{Overall Objective}
\label{sec:overall_objective}
In addition to the two sKLD constraint loss terms mentioned in the above subsection, we also use the standard cross-entropy (CE) loss:
\begin{equation}
    \mathcal{L}_\text{CE}=-\frac{1}{N+M}\sum^{N+M}_{i=1}\sum^{C^l+C^u}_{k=1}\bm{y}_i(k)\log\bm{p}_i(k),
\end{equation}
where $\bm{y}_i(k)$ indicates the zero-padded ground truth label of image $\bm{x}_i$ if $\bm{x}_i$ is a labelled image, otherwise it is the zero-padded pseudo label of $\bm{x}_i$, predicted by the Sinkhorn-Knopp algorithm like UNO, and $\bm{p}_i$ is the predicted probability distribution corresponding to $\bm{x}_i$. In summary, with the purpose of maximizing the inter-class sKLD and minimizing the intra-class sKLD at the same time, the overall objective function in our model is
\begin{equation}
    \mathcal{L}=\mathcal{L}_\text{CE}-\alpha\mathcal{L}_\text{inter-class}+\beta\mathcal{L}_\text{intra-class},
\end{equation}
where $\alpha,\beta>0$ are two hyperparameters of the two sKLD loss terms.

\begin{table}[!t]\small
\centering
\tabcolsep=2pt
\caption{Details of dataset splits used in the experiments.}
\begin{tabular}{l*{4}{c}}
    \toprule
    \multirow{2}{*}{Dataset split} & \multicolumn{2}{c}{Labelled} & \multicolumn{2}{c}{Unlabelled} \\
    \cmidrule(lr){2-3}\cmidrule(lr){4-5}
    & \#Images & \#Classes & \#Images & \#Classes \\
    \midrule
    CIFAR10 & 25K & 5 & 25K & 5 \\
    CIFAR100-20 & 40K & 80 & 10K & 20 \\
    CIFAR100-50 & 25K & 50 & 25K & 50 \\
    ImageNet & 1.25M & 882 & $\approx$30K & 30 \\
    \bottomrule
\end{tabular}
\label{dataset}
\end{table}

\begin{table*}[!t]\small
\centering
\caption{Ablation study of our method on three dataset splits. Results are reported in ACC (\%), NMI and ARI that are averaged over 5 runs. ``Inter-class'' stands for the inter-class sKLD loss term and ``Intra-class'' is short for the intra-class sKLD. \textbf{Best} results are highlighted in each column.}
\begin{tabular}{*{12}{c}}
    \toprule
    \multirow{2}{*}{\#} & \multirow{2}{*}{Inter-class} & \multirow{2}{*}{Intra-class} & \multicolumn{3}{c}{CIFAR10} & \multicolumn{3}{c}{CIFAR100-20} & \multicolumn{3}{c}{CIFAR100-50} \\
    \cmidrule(lr){4-6}\cmidrule(lr){7-9}\cmidrule(lr){10-12}
     & & & ACC & NMI & ARI & ACC & NMI & ARI & ACC & NMI & ARI \\
    \midrule
    1 &\XSolidBrush &\XSolidBrush & 93.25 & 0.8717 & 0.8565 & 90.54 & 0.8475 & 0.8123 & 62.27 & 0.6781 & 0.4688 \\
    2 &\Checkmark   &\XSolidBrush & 99.07 & 0.9644 & 0.9770 & 92.17 & 0.8695 & 0.8448 & 64.31 & 0.7046 & 0.5102 \\
    3 &\XSolidBrush &\Checkmark   & 93.60 & 0.8753 & 0.8624 & 91.13 & 0.8506 & 0.8193 & 63.35 & 0.6800 & 0.4729 \\
    4 &\Checkmark   & \Checkmark & \textbf{99.11} & \textbf{0.9657} & \textbf{0.9780} & \textbf{92.48} & \textbf{0.8727} & \textbf{0.8508} & \textbf{65.85} & \textbf{0.7106} & \textbf{0.5238} \\
    \bottomrule
\end{tabular}
\label{ablation}
\end{table*}

\section{Experiments}
\subsection{Experimental Setup}
\textbf{Datasets.}
We perform our experiments on three benchmark datasets that are widely-used in the field of NCD, including CIFAR10~\cite{CIFAR}, CIFAR100~\cite{CIFAR} and ImageNet~\cite{ImageNet}. Following the literature of NCD~\cite{RS_ICLR2020, NCL_CVPR2021, ComEx_CVPR2022}, each dataset is further divided into two parts for the NCD setting: a labelled subset and an unlabelled subset. Also, we assume the number of classes in the unlabelled subset is known a priori. In addition, according to the number of classes contained in the subsets, there are four different dataset splits: CIFAR10, CIFAR100-20, CIFAR100-50 and ImageNet, where CIFAR100-50 is first introduced in UNO~\cite{UNO_ICCV2021} as a more challenging evaluation than CIFAR100-20. More details of the dataset splits are shown in~\cref{dataset}.

\textbf{Evaluation Metrics.}
Following the evaluation protocols used in the literature~\cite{UNO_ICCV2021, ComEx_CVPR2022}, we also conduct our experiments using both task-aware and task-agnostic evaluation protocols. In the task-aware protocol, we know the task information in advance when evaluating a method, that is, we know whether an image is from the labelled subset or the unlabelled one. In contrast, in the task-agnostic protocol, we have no idea about such information. Following the literature, we use the average clustering accuracy ($\mathcal{ACC}$) to evaluate the performance of our method, defined as
\begin{equation}
    \mathcal{ACC} = \max_{\text{perm} \in P}\frac{1}{N}\sum^N_{i=1}\mathbbm{1}\{y_i=\text{perm}(\hat{y}_i)\},
\end{equation}
where $y_i$ and $\hat{y}_i$ are the ground-truth label and clustering assignment of a test sample $x^u_i \in D^u$, respectively. $P$ is the set of all permutations and the optimal permutation can be calculated by the Hungarian algorithm~\cite{Hungarian}.

\textbf{Implementation Details.}
For a fair comparison with the existing methods, we use ResNet-18~\cite{background_ResNet} as the backbone and two steps to train our model. We first pre-train the model for 200 epochs on the labelled subset, and then fine-tune the pre-trained model for 500 epochs to discover novel classes. For all experiments, we fix the batch size to 512. Note that the proposed inter-class and intra-class symmetric Kullback-Leibler divergence (sKLD) terms are only used in the training phase and will be discarded in the test phase. Specifically, for the inter-class sKLD, we set the weight $\alpha$ to 0.02 for CIFAR100-20 and 0.05 for the other three dataset splits. As for the intra-class sKLD, we fix $\beta=0.01$ on all benchmarks. Following the literature~\cite{background_DeepCluster,background_IIC,UNO_ICCV2021,ComEx_CVPR2022}, we use multi-head clustering and overclustering to boost the clustering performance. Besides, we apply multi-crop strategy and common data augmentations (\eg, random crop, flip, color jittering, and grey-scale) to generate transformed samples. In addition, we conduct all the experiments on the Tesla V100 GPUs.

\subsection{Ablation Study}
\label{sec:ablation}
\textbf{Symmetric KLD vs. MSE for Intra-class Constraint.}
As mentioned above, many previous works~\cite{NCL_CVPR2021,UNO_ICCV2021,ComEx_CVPR2022} have used data augmentation (\eg, rotation, flipping and random crop) to generate randomly transformed samples for obtaining pseudo labels. To further ensure consistent predictions between different augmentations of the same image, \ie, intra-class constraint, RS~\cite{RS_ICLR2020} and DualRank~\cite{DualRank_NeurIPS2021} introduce the mean squared error (MSE) as a regularization term, which is a common operation in semi-supervised learning. However, we find that the MSE-based intra-class constraint does not perform well in practice. Therefore, we employ an symmetric KLD measure instead of the common MSE to constrain such an intra-class consistency.

To verify the effectiveness of symmetric KLD, we compare the proposed intra-class sKLD with MSE and report the results averaged over $5$ runs in~\cref{sKLD_vs_MSE}. Note that the hyperparameters of both sKLD and MSE have been well selected through grid search. From the results, we can see that the MSE constraint can only slightly improve the performance of Baseline on CIFAR100-20 and CIFAR100-50, and even somewhat damages the performance on CIFAR10. On the contrary, the proposed sKLD constraint could consistently boost the performance of Baseline on all benchmarks. The intuitive reason is that the semantics of different augmentations (especially strong data augmentations) of the same image has been changed to some extent. Therefore, we should not strictly make their predictions be the same did as the MSE constraint. In contrast, the slacker sKLD constraint just making the predictions be similar will be more beneficial to the final clustering.

\begin{table}[!t]\small
\centering
\tabcolsep=5pt
\caption{Ablation study on the intra-class constraint on CIFAR-10 and CIFAR-100. ``sKLD'' and ``MSE'' denote the intra-class sKLD and MSE consistency regularization term, respectively. Results are reported in ACC (\%) averaged over 5 runs. \textbf{Best} results are highlighted in each column.}
\begin{tabular}{l c c c}
    \toprule
    Method & CIFAR10 & CIFAR100-20 & CIFAR100-50 \\
    \midrule
    Baseline & 93.25 & 90.54 & 62.27 \\
    + MSE    & 93.11 & 90.96 & 62.47 \\
    + sKLD   & \textbf{93.60}  & \textbf{91.13} & \textbf{63.35} \\
    \bottomrule
\end{tabular}
\label{sKLD_vs_MSE}
\end{table}

\begin{table*}[!t] \small
\centering
  \tabcolsep=8pt
\caption{Comparison with state-of-the-art methods on the training split of the unlabelled subset, using task-aware evaluation protocol. Results are reported in clustering accuracy (\%) in the form of mean and standard deviation (averaged over 5 runs except for ImageNet). \textbf{Best} results are highlighted in each column. $^\dagger$Our reproduced result.}
\begin{tabular}{l c c c c c c}
    \toprule
    Method & Venue & Type & CIFAR10 & CIFAR100-20 & CIFAR100-50 & ImageNet\\
    \midrule
    $k$-means & Classic & - & 72.5$\pm$0.0 & 56.3$\pm$1.7 & 28.3$\pm$0.7 & 71.9\\
    KCL & ICLR'18 & Two-stage & 72.3$\pm$0.2 & 42.1$\pm$1.8 & – & 73.8\\
    MCL & ICLR'19 & Two-stage & 70.9$\pm$0.1 & 21.5$\pm$2.3 & – & 74.4\\
    DTC & ICCV'19 & Two-stage & 88.7$\pm$0.3 & 67.3$\pm$1.2 & 35.9$\pm$1.0 & 78.3\\
    \midrule
    RS & ICLR'20 & Single-stage & 90.4$\pm$0.5 & 73.2$\pm$2.1 & 39.2$\pm$2.3 & 82.5\\
    RS+ & ICLR'20 & Single-stage & 91.7$\pm$0.9 & 75.2$\pm$4.2 & 44.1$\pm$3.7 & 82.5\\
    OpenMix & CVPR'21 & Single-stage & 95.3 & – & – & 85.7\\
    NCL & CVPR'21 & Single-stage & 93.4$\pm$0.5 & 86.6$\pm$0.4 & – & 90.7\\
    Joint & ICCV'21 & Single-stage & 93.4$\pm$0.6 & 76.4$\pm$2.8 & – & 86.7\\
    UNOv1 & ICCV'21 & Single-stage & 96.1$\pm$0.5 & 85.0$\pm$0.6 & 52.9$\pm$1.4 & 90.6\\
    UNOv2$^\dagger$ & ICCV'21 & Single-stage & 93.3$\pm$0.4 & 90.5$\pm$0.7 & 62.3$\pm$1.4 & 90.7\\
    DualRank & NeurIPS'21 & Single-stage & 91.6$\pm$0.6 & 75.3$\pm$2.3 & – & 88.9\\
    ComEx & CVPR'22 & Single-stage & 93.6$\pm$0.3 & 85.7$\pm$0.7 & 53.4$\pm$1.3 & 90.9\\
    \midrule
    \textbf{IIC} (Ours) & CVPR'23 & Single-stage & \textbf{99.1$\pm$0.0} & \textbf{92.4$\pm$0.2} & \textbf{65.8$\pm$0.9} & \textbf{91.9}\\
    \bottomrule
\end{tabular}
\label{SOTA_aware}
\end{table*}

\begin{table*}[!t] \small
\centering
\tabcolsep=10pt
\caption{Comparison with state-of-the-art methods on CIFAR-10 and CIFAR-100, using task-agnostic evaluation protocol. Results are reported averaged over 5 runs in classification accuracy (\%) and clustering accuracy (\%) on test split of the labelled and unlabelled subsets, respectively. \textbf{Best} results are highlighted in each column.}
\begin{tabular}{l *{9}{c}}
    \toprule
    \multirow{2}{*}{Method} & \multicolumn{3}{c}{CIFAR10} & \multicolumn{3}{c}{CIFAR100-20} & \multicolumn{3}{c}{CIFAR100-50} \\
    \cmidrule(lr){2-4}\cmidrule(lr){5-7}\cmidrule(lr){8-10}
     & Label & Unlabel & All & Label & Unlabel & All & Label & Unlabel & All \\
    \midrule
    KCL & 79.4 & 60.1 & 69.8 & 23.4 & 29.4 & 24.6 & – & – & -\\
    MCL & 81.4 & 64.8 & 73.1 & 18.2 & 18.0 & 18.2 & – & – & -\\
    DTC & 58.7 & 78.6 & 68.7 & 47.6 & 49.1 & 47.9 & 30.2 & 34.7 & 32.5\\
    RS+ & 90.6 & 88.8 & 89.7 & 71.2 & 56.8 & 68.3 & 69.7 & 40.9 & 55.3\\
    UNOv1 & 93.5 & 93.3 & 93.4 & 73.2 & 73.1 & 73.2 & 71.5 & 50.7 & 61.1\\
    ComEx & 95.0 & 92.6 & 93.8 & 75.2 & 77.3 & 75.6 & \textbf{75.3} & 53.5 & 64.4\\
    \midrule
    \textbf{IIC} (Ours) & \textbf{96.0} & \textbf{97.2} & \textbf{96.6} & \textbf{75.9} & \textbf{78.4} & \textbf{77.2} & 75.1 & \textbf{61.0} & \textbf{68.1}\\
    \bottomrule
\end{tabular}
\label{SOTA_agnostic}
\end{table*}

\begin{figure*}[t]
  \centering
  \begin{subfigure}{0.325\linewidth}
    \includegraphics[width=0.7\linewidth]{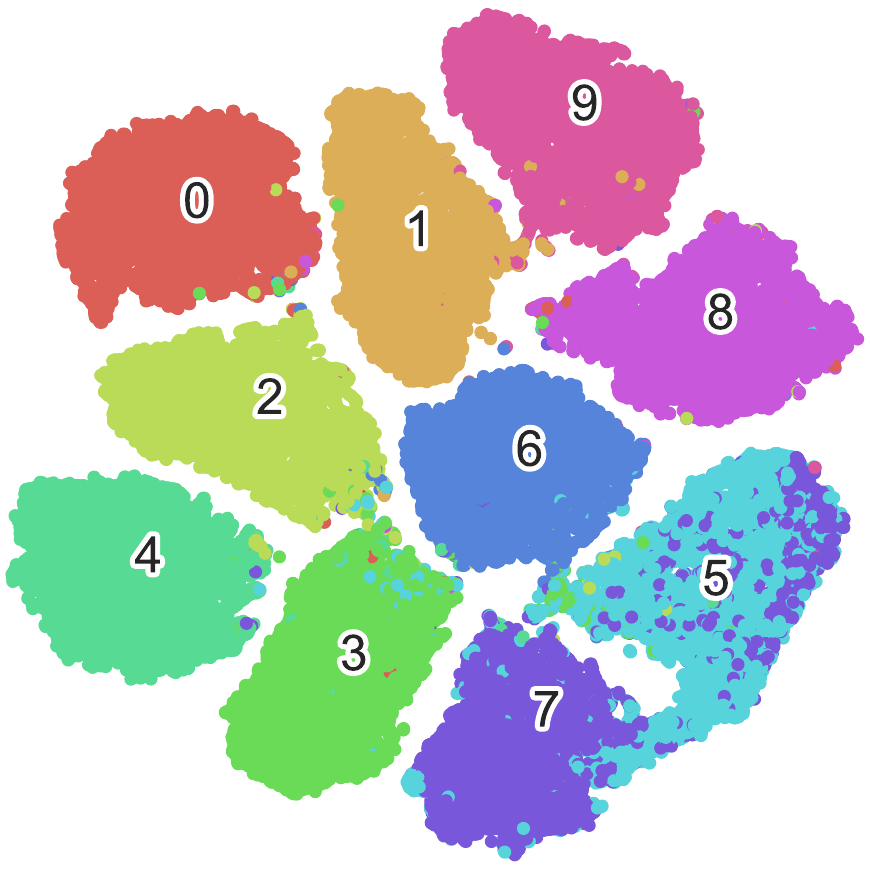}
    \caption{UNOv2}
  \end{subfigure}
  \hfill
  \begin{subfigure}{0.325\linewidth}
    \includegraphics[width=0.7\linewidth]{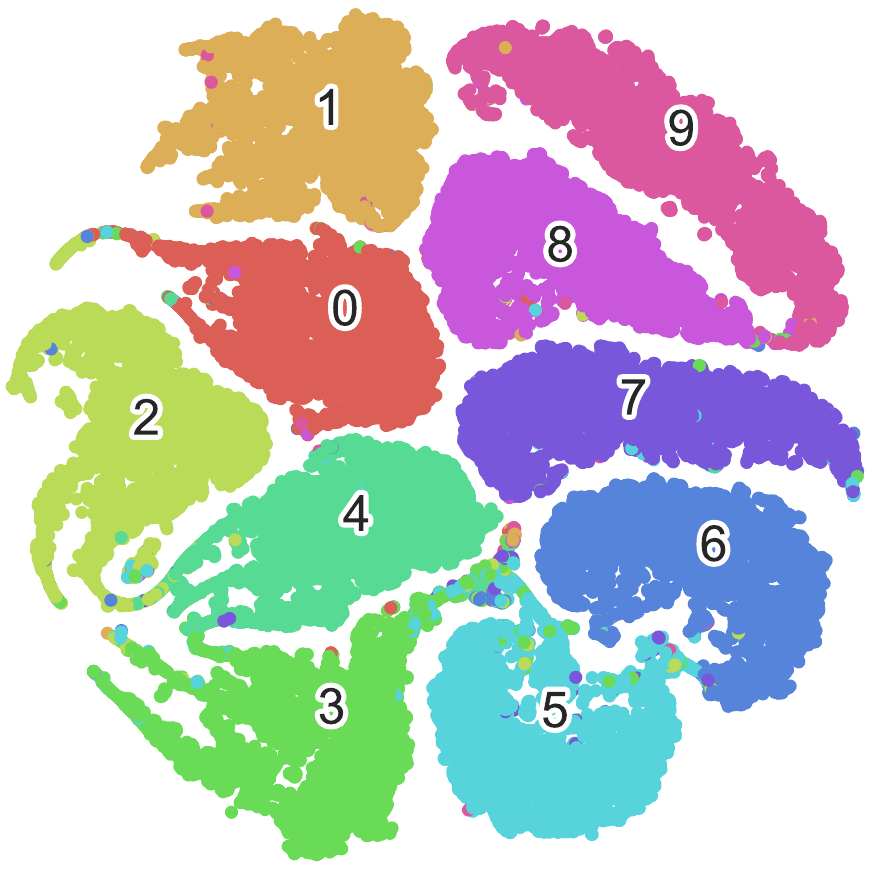}
    \caption{Ours}
  \end{subfigure}
  \hfill
  \begin{subfigure}{0.325\linewidth}
    \includegraphics[width=0.7\linewidth]{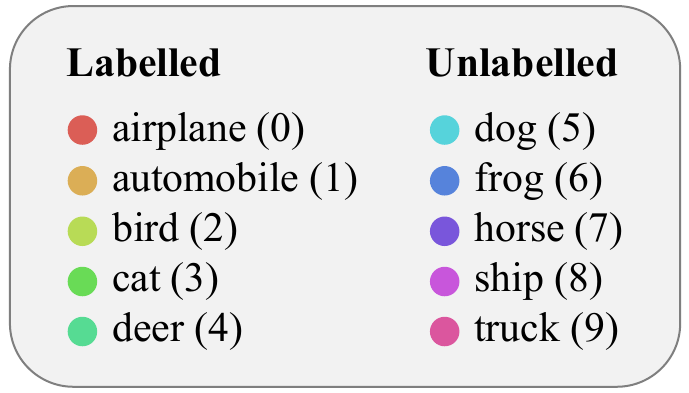}
  \end{subfigure}
  \caption{t-SNE visualization on CIFAR10. We compare our method with the reproduced UNOv2. Best viewed in color, each color indicates a different class and we also mark the index corresponding to each class.}
  \label{visualization}
\end{figure*}

\textbf{Symmetric KLD Constraints.}
In this subsection, we perform ablation study experiments on three dataset splits, including CIFAR10, CIFAR100-20 and CIFAR100-50, to evaluate the effectiveness of the proposed \textit{inter-class sKLD} and \textit{intra-class sKLD} components. Besides the ACC, we also report two extra evaluation criteria that are commonly used in clustering: \ie, normalized mutual information (NMI)~\cite{NMI} and adjusted rand index (ARI)~\cite{ARI}. MNI and ARI are generally used to measure the similarity between the clustering results and the ground-truth distributions. The closer the criteria are to 1, the better the clustering effect. The results on ablation study are shown in~\cref{ablation}.

Considering that the trends of three evaluation criteria (\ie, ACC, NMI and ARI) on all dataset splits are similar in~\cref{ablation}, we will mainly take the ACC as an example for discussion. Specifically, as seen, the proposed inter-class sKLD alone can significantly improve the ACC of the baseline (\ie, UNO) on all benchmarks, especially on CIFAR10 where the performance improvement is over $5.82\%$. Although the proposed intra-class sKLD alone do not distinctly boost the performance on CIFAR10, it tangibly gains $0.59\%$ and $1.08\%$ ACC improvements over the baseline on CIFAR100-20 and CIFAR100-50, respectively.

Notably, when combining the inter-class sKLD and intra-class sKLD together, we observe that the results on all three dataset splits, especially on CIFAR100-20 and CIFAR100-50, are further improved, and the improvements are remarkable and stable. It is worth mentioning that because the CIFAR10 dataset is somewhat simple, the inter-class sKLD alone has made the ACC reach the saturation point and thus adding the additional intra-class sKLD does not bring distinct improvements. As expected, on the more challenging benchmarks, such as CIFAR100-20 and CIFAR100-50, integrating the inter-class sKLD with intra-class sKLD together can further obtains $0.31\%$ and $1.54\%$ ACC improvements over using the inter-class sKLD alone, respectively.

In summary, we are able to conclude that both inter-class and intra-class sKLD constraints are effective and each of them can consistently boost the performance of the baseline. In particular, it is clearly that the inter-class sKLD constraint alone can gain more improvements than the intra-class sKLD constraint alone. This is because the purpose of inter-class sKLD is to use prior information of NCD, \ie, the unlabelled classes are naturally disjoint with the labelled classes, to learn a more discriminative embedding space. Differently, the intra-class sKLD is more like a consistency regularization term that is used to explicitly constrain samples and their augmentations. As discussed above, the intra-class sKLD, as a slacker regularization term than the MSE-based one, will bring a greater positive impact on the more difficult datasets or their splits.

\subsection{Comparison with State of the Arts}
We compare our proposed method with the current state-of-the-art methods in the field of NCD, including KCL~\cite{KCL_ICLR2018}, MCL~\cite{MCL_ICLR2019}, DTC~\cite{DTC_ICCV2019}, RS~\cite{RS_ICLR2020}, RS+ (RS with incremental classifier)~\cite{RS_ICLR2020}, OpenMix~\cite{OpenMix_CVPR2021}, DualRank~\cite{DualRank_NeurIPS2021}, Joint~\cite{Joint_ICCV2021}, NCL~\cite{NCL_CVPR2021}, UNO~\cite{UNO_ICCV2021} and ComEx~\cite{ComEx_CVPR2022} besides the classical $k$-means~\cite{k-means}. We report the results on four popular benchmarks in~\cref{SOTA_aware} and~\cref{SOTA_agnostic} using task-aware and task-agnostic evaluation protocols, respectively. It is worth noting that UNO has two versions, \ie, UNOv1 and UNOv2. The results of UNOv1 are quoted from the original paper, while we report our reproduced results of UNOv2 using the officially released code.

In~\cref{SOTA_aware}, we report the average clustering accuracy (\ie, ACC) of all methods on the training split of the unlabelled set, using a task-aware evaluation protocol (commonly used in the literature). As we can see, our proposed method outperforms the current works on all benchmarks. In particular, our method surpasses the latest ComEx by a significant margin, such as, gaining $5.5\%$ improvements on CIFAR10 and $12.4\%$ improvements on CIFAR100-50, respectively. Compared with the reproduced UNOv2, our method still has a certain improvement, \ie, about $5.8\%$ improvements on CIFAR10, $1.9\%$ improvements on CIFAR100-20, $3.5\%$ improvements on CIFAR100-50 and $1.2\%$ improvements on ImageNet. Note that the standard deviations of our method are also smaller than the previous works, indicating that the proposed method does well in stability.

We also care about the performance on test split of the labelled and unlabelled sets with a task-agnostic evaluation protocol and report the results in terms of the classification accuracy and clustering accuracy in~\cref{SOTA_agnostic}. As seen, our method outperforms the state-of-the-art methods on both labelled and unlabelled test set of all benchmarks except for CIFAR100-50, which successfully demonstrate its effectiveness and robustness. Specifically, for CIFAR10 and CIFAR100-20, our method not only is superior to the related works on the labelled set, but also obtains greater performance improvements on the unlabelled set, achieving the excellent comprehensive results. Moreover, for CIFAR100-50, our method is very close to the latest ComEx on the labelled subset (\ie, with only $0.2\%$ difference) and gains considerable improvements over ComEx on the unlabelled classes by $7.5\%$. Therefore, the results compared with the current state-of-the-art methods in~\cref{SOTA_aware} and~\cref{SOTA_agnostic} show that the proposed inter- and intra-class sKLDs indeed help to discriminate labelled points from unlabelled ones in the embedding space, both of which are simple yet effective.

\begin{figure}[!t]
  \centering
  \begin{subfigure}{0.49\columnwidth}
    \includegraphics[width=0.75\linewidth]{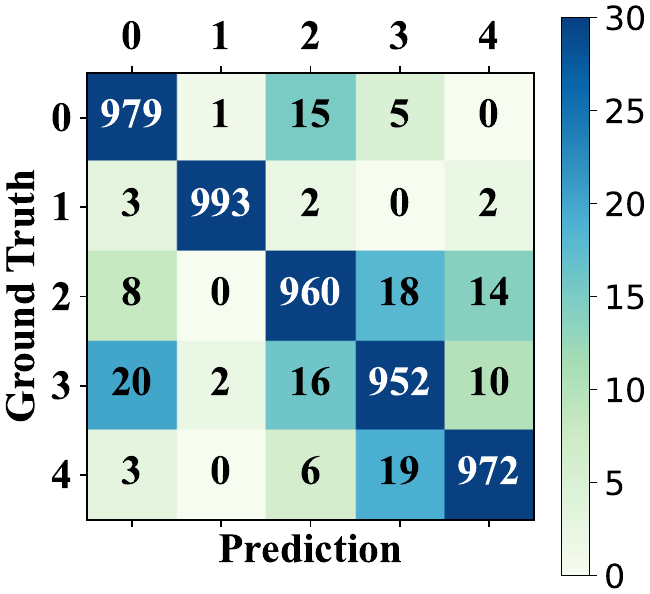}
    \caption{UNOv2 on labelled subset}
    \label{baseline_labelled_aware}
  \end{subfigure}
  \hfill
  \begin{subfigure}{0.49\columnwidth}
    \includegraphics[width=0.75\linewidth]{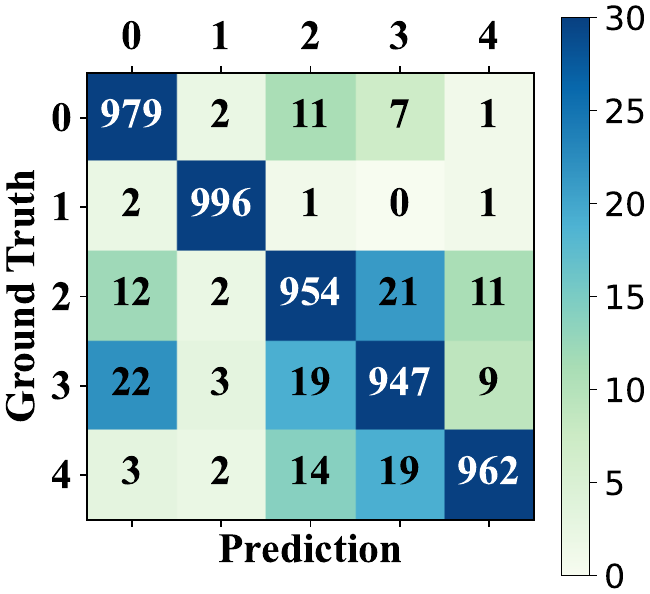}
    \caption{UNOv2+ on labelled subset}
    \label{inter_labelled_aware}
  \end{subfigure}
  
  \begin{subfigure}{0.49\columnwidth}
    \includegraphics[width=0.75\linewidth]{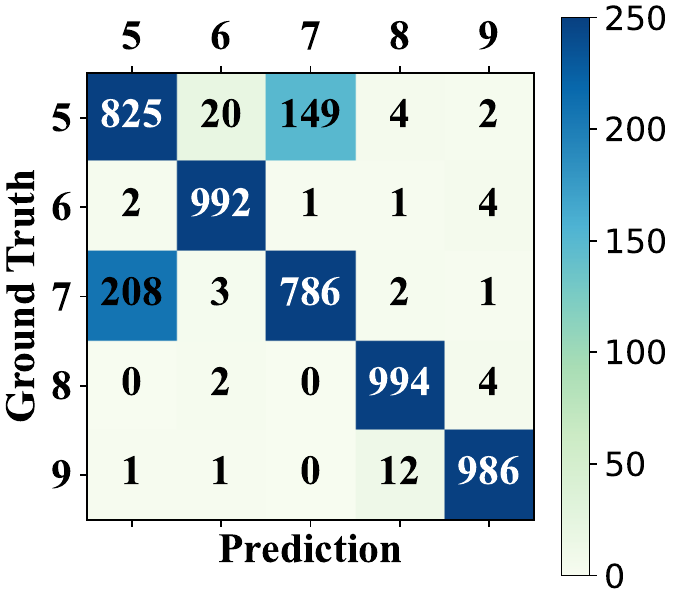}
    \caption{UNOv2 on unlabelled subset}
    \label{baseline_unlabelled_aware}
  \end{subfigure}
  \hfill
  \begin{subfigure}{0.49\columnwidth}
    \includegraphics[width=0.75\linewidth]{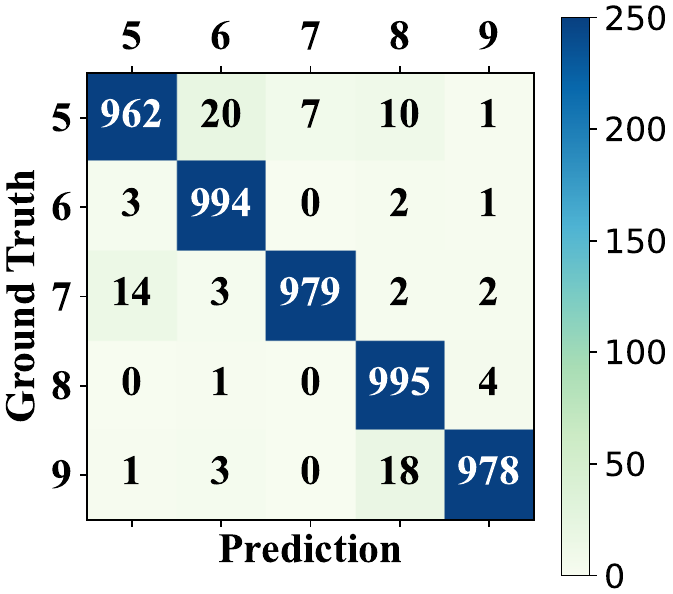}
    \caption{UNOv2+ on unlabelled subset}
    \label{inter_unlabelled_aware}
  \end{subfigure}
  \caption{Confusion matrices of the reproduced UNO without (UNOv2) and with (UNOv2+) the inter-class sKLD constraint on CIFAR10, using the task-aware evaluation protocol. The two models are evaluated on test split of the labelled subset (\cref{baseline_labelled_aware}, \cref{inter_labelled_aware}) and unlabelled subset (\cref{baseline_unlabelled_aware}, \cref{inter_unlabelled_aware}), respectively.}
  \label{confusion_matrix_aware}
\end{figure}

\subsection{Visualization}
\textbf{t-SNE Visualization.}
To better demonstrate the proposed method, we illustrate the t-SNE~\cite{t-SNE} visualization for all classes on CIFAR10, where the first five classes are labelled (\ie, airplane, automobile, bird, cat and deer) and the remaining five classes are unlabelled (\ie, dog, frog, horse, ship and truck). In~\cref{visualization}, we compare our method with the reproduced UNOv2~\cite{UNO_ICCV2021} through its officially released code, which is the current state-of-the-art method. Furthermore, to ensure a fair comparison, we follow the approach used in UNOv2 and use the concatenated logits $\bm{l}$ from both classification heads $h$ and $g$ to perform the t-SNE visualization.

As we can see, in general, two methods can well accomplish the NCD task, and there are only differences between them in the clustering ability on the certain classes. To be specific, for some labelled classes (\eg, bird (2), cat (3) and deer (4)), UNOv2 shows more compact clustering results. Nevertheless, our method distinguishes better on the unlabelled classes (\eg, dog (5) and horse (7)), demonstrating that our method can learn more discriminative feature representations with better generalization. To further show this point, we draw attention to two unlabelled classes, \ie, dog (5) and horse (7). We can see that UNOv2 completely confuses these two classes, whose distributions are basically overlapped with each other in~\cref{visualization}. We surmise the reason is that the poses of these two classes are visually similar, resulting in the poor clustering performance of UNOv2. In contrast, our method can clearly distinguish these two classes, because the proposed inter-class constraint can effectively separate different classes and the proposed intra-class constraint can compact the distribution of a class.

\textbf{Confusion Matrix.}
Like RS~\cite{RS_ICLR2020} and UNO~\cite{UNO_ICCV2021}, we transform the original clustering task into a classification task for the NCD problem on the premise of knowing the number of categories of the unlabelled data. Therefore, we can validate the proposed method by the confusion matrix, which is commonly used in generic classification tasks. According to the ablation study in~\cref{sec:ablation}, it has been concluded that the proposed inter-class sKLD constraint plays a key role in our method. Therefore, it will be interesting to visually explain the effect of such a constraint. To be specific, we add the inter-class sKLD constraint into UNOv2 to obtain a new variant UNOv2+. Next, we establish the confusion matrices between the two models before and after using such an inter-class sKLD constraint.

\begin{figure}[!t]
  \centering
  \begin{subfigure}{0.48\linewidth}
    \includegraphics[width=1\linewidth]{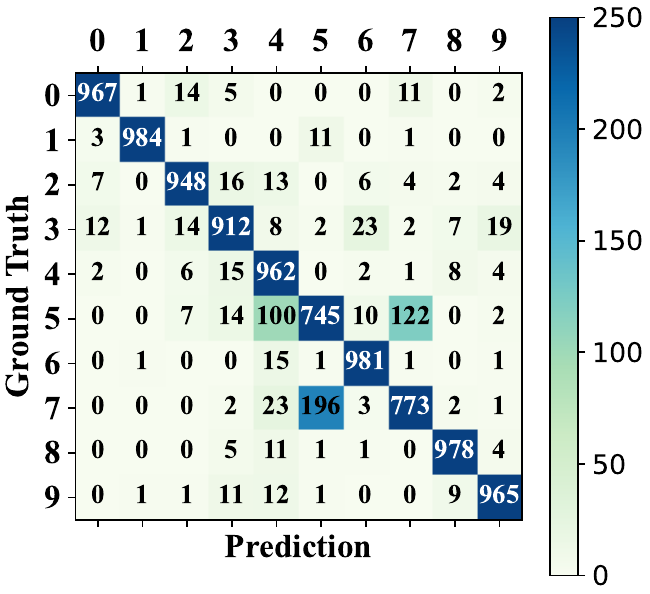}
    \caption{UNOv2}
  \end{subfigure}
  \hfill
  \begin{subfigure}{0.48\linewidth}
    \includegraphics[width=1\linewidth]{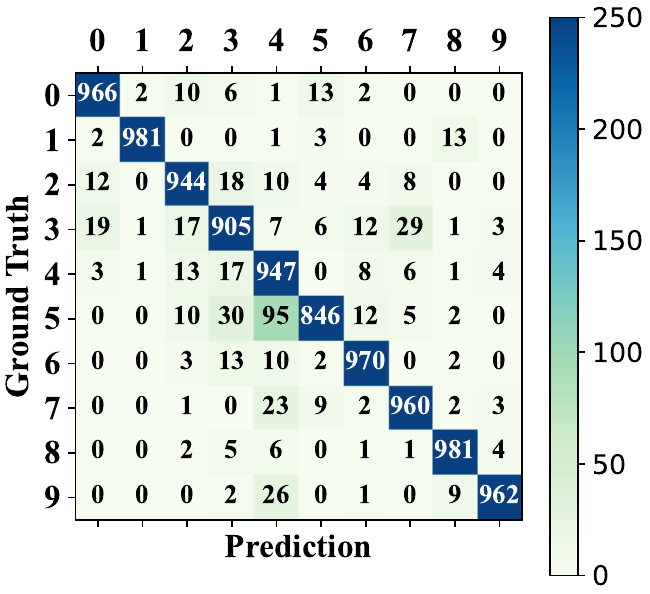}
    \caption{UNOv2+}
  \end{subfigure}
  \caption{Confusion matrices of the reproduced UNO without (UNOv2) and with (UNOv2+) the inter-class sKLD constraint on the test split of CIFAR10, using the task-agnostic evaluation protocol, where each image is predicted into one of the 10 classes.}
  \label{confusion_matrix_agnostic}
\end{figure}

The results are reported on the test split of CIFAR10 in~\cref{confusion_matrix_aware} and~\cref{confusion_matrix_agnostic} using task-aware and task-agnostic evaluation protocols, respectively. In~\cref{confusion_matrix_aware}, we compare the performance between UNOv2 and UNOv2+ on the labelled and unlabelled subsets. Specifically, we observe that the misclassification of UNOv2 is almost unchanged on the labelled subset after adding the inter-class sKLD constraint, but UNOv2+ obtains significant improvements for the unlabelled classes. With the help of the inter-class sKLD constraint, UNOv2+ will no longer confuse dog ($5$) and horse ($7$). In addition, \cref{confusion_matrix_agnostic} demonstrates that UNOv2+ can prevent misclassifying a labelled image into unlabelled classes or vice versa. These results show that the proposed inter-class sKLD constraint can effectively distinguish unlabelled data from labelled data to help discovering novel classes, and can be easily added into other single-stage based NCD methods as a plug-and-play loss term.

\section{Conclusion}
In this paper, we propose to model both inter-class and intra-class constraints built on the symmetric Kullback-Leibler divergence (sKLD) for novel class discovery (NCD). We conduct extensive experiments on four popular benchmarks and show that our method could outperform the existing state-of-the-art methods by a large margin. From the experimental results, we have the following findings: (1) making use of the disjoint characteristic between the labelled and unlabelled classes, \ie, constraining an inter-class constraint, is important and effective for NCD; (2) using a sKLD measure instead of the MSE for constraining the intra-class constraint is reasonable and beneficial to NCD.

\noindent \textbf{Acknowledgements.} \quad This work is supported in part by the National Natural Science Foundation of China (62106100, 62192783, 62276128), Jiangsu Natural Science Foundation (BK20221441), CAAI-Huawei MindSpore Open Fund, the Collaborative Innovation Center of Novel Software Technology and Industrialization, and Jiangsu Provincial Double-Innovation Doctor Program (JSSCBS20210021).

{\small
\bibliographystyle{ieee_fullname}
\bibliography{egbib}
}

\appendix
\section*{Appendix}
\renewcommand\thetable{\Alph{section}}

\section{Varying the Number of Clusters}
To verify the effectiveness of the proposed method in more complex scenarios, we consider dataset splits with multiple different levels of difficulty and conduct additional experiments in~\cref{increasing_clusters}. In the similar way of dividing CIFAR100 into CIFAR100-20 and CIFAR100-50 in the main paper, we adjust the number of classes contained in the labelled and unlabelled subsets to obtain various different dataset splits, such as CIFAR100-30 and CIFAR100-80.

As shown in~\cref{increasing_clusters}, we compare the proposed method with UNOv2~\cite{UNO_ICCV2021} using an increasing number of unlabelled classes on CIFAR100 and report the corresponding task-aware clustering accuracy. Despite the gradual decrease in performance of both methods with increasing difficulty of the NCD task, our method consistently achieves superior results compared with UNOv2, with a minimum improvement of $1.2\%$ in ACC. In this sense, we are able to demonstrate that the proposed method can achieve solid performance even facing with more challenging NCD tasks.

\begin{table*}[!t]
\centering
\caption{Experimental results with an increasing number of unlabelled classes on CIFAR100, using the task-aware evaluation protocol. Results are reported in clustering accuracy (\%) in form of mean and standard deviation (averaged over 3 runs). For simplicity, we fix the hyperparameters $\alpha=0.05$ and $\beta=0.01$ for the inter-class constraint and the intra-class constraint, respectively. \textbf{Best} results are highlighted in each column. $^\dagger$Our reproduced result.}
\begin{tabular}{l *{7}{c}}
    \toprule
    \multirow{2}{*}{Method} & \multicolumn{7}{c}{\#Unlabelled classes} \\
    \cmidrule(lr){2-8}
     & 20 & 30 & 40 & 50 & 60 & 70 & 80 \\
    \midrule
    UNOv2$^\dagger$~\cite{UNO_ICCV2021} & 90.5$\pm$0.7 & 76.7$\pm$2.0 & 66.8$\pm$1.7 & 62.3$\pm$1.4 & 57.9$\pm$0.9 & 56.0$\pm$1.1 & 56.2$\pm$1.0 \\
    \textbf{IIC} (Ours) & \textbf{92.4$\pm$0.2}  & \textbf{83.8$\pm$1.6} & \textbf{71.1$\pm$1.4} & \textbf{65.8$\pm$0.9} & \textbf{60.6$\pm$0.4} & \textbf{57.7$\pm$1.0} & \textbf{57.4$\pm$0.9} \\
    \bottomrule
\end{tabular}
\label{increasing_clusters}
\end{table*}

\section{Unknown Number of Novel Classes}
The results reported in the main paper are based on a common assumption, \ie, the number of novel classes $C^u$ is known a priori, in NCD. However, in many real applications, it is somewhat difficult to obtain this priori in advance. That is to say, when facing with a real unlabelled dataset, we may cannot foretell the number of clusters accurately. To address this issue, many previous NCD works relax this restriction by using an estimation algorithm proposed in DTC~\cite{DTC_ICCV2019} to estimate the number of unlabelled classes before discovering novel classes. For example, DualRank~\cite{DualRank_NeurIPS2021} and UNO~\cite{UNO_ICCV2021} are both in this way. To be specific, DTC splits a probe subset from the labelled dataset, and then runs a semi-supervised $k$-means algorithm multiple times on the union of the probe subset and the unlabelled dataset by varying the number of clusters. The optimal number of novel classes is finally obtained by optimizing the cluster quality indices on the probe subset and unlabelled dataset.

In order to further demonstrate the effectiveness of our method in a more realistic setting, we first use the aforementioned estimation algorithm to find an estimation of novel classes on the CIFAR100-20 dataset split, denoted by $\hat{C^u}$. Next, we compare our proposed method with UNOv2 using the estimated number of novel classes and report the final clustering results in~\cref{unknown_number}. It can be found that our method consistently outperforms UNOv2 and other comparison methods by a large margin and our method can achieve more stable performance (\ie, with smaller standard deviations), when the ground truth number of novel classes is unknown.

\begin{table}[!t]
\centering
\caption{Experimental results with the estimated number of novel classes on CIFAR100-20, using the task-aware evaluation protocol. We also report the result with the true number of unlabelled classes. Results are reported in clustering accuracy (\%) in form of mean and standard deviation (averaged over 3 runs). For simplicity, we fix $\alpha=0.05$ and $\beta=0.01$. \textbf{Best} results are highlighted in each column. $^\dagger$Our reproduced result.}
\begin{tabular}{l *{2}{c}}
    \toprule
    \multirow{2}{*}{Method} & \multicolumn{2}{c}{Class number} \\
    \cmidrule(lr){2-3}
     & $C^u=20$ & $\hat{C^u}=23$ \\
    \midrule
    DTC~\cite{DTC_ICCV2019} & 67.3$\pm$1.2 & 64.3 \\
    RS~\cite{RS_ICLR2020} & 73.2$\pm$2.1 & 70.5 \\
    RS+~\cite{RS_ICLR2020} & 75.2$\pm$4.2 & 71.2 \\
    UNOv1~\cite{UNO_ICCV2021} & 85.0$\pm$0.6 & 75.1 \\
    UNOv2$^\dagger$~\cite{UNO_ICCV2021} & 90.5$\pm$0.7 & 71.5$\pm$1.8 \\
    \midrule
    \textbf{IIC} (Ours) & \textbf{92.4$\pm$0.2} & \textbf{85.1$\pm$0.9} \\
    \bottomrule
\end{tabular}
\label{unknown_number}
\end{table}

\section{Speeding up the Computation of Inter-Class Constraint}
From the experimental results and analyses in the main paper, it can be concluded that our proposed symmetric Kullback-Leibler divergence (sKLD) based constraints are simple yet effective. However, the calculation speed will be very slow if directly using the equations mentioned in~\cref{sec:skld}, especially when the number of samples is large. It can be found that the reason for the enormous time cost is that calculating the inter-class sKLD loss term requires nested loops with the division operation (\cref{inter_1} - \cref{inter_4} in main paper). Therefore, we have to make additional designs for the objective of inter-class constraint to speed up the computation. Taking~\cref{inter_2} as an example, we can simply rewrite the fraction into a subtraction formula as
\begin{equation}\small
\label{new_kld}
\begin{split}
    D_{\rm KL}(\bm{p}^l_i||\bm{p}^u_j) &= \sum^{C^l+C^u}_{k=1}\bm{p}^l_i(k)\log\frac{\bm{p}^l_i(k)}{\bm{p}^u_j(k)} \\
    &= \sum^{C^l+C^u}_{k=1}\bm{p}^l_i(k)(\log\bm{p}^l_i(k)-\log\bm{p}^u_j(k)) \\
    &= \sum^{C^l+C^u}_{k=1}\bm{p}^l_i(k)\log\bm{p}^l_i(k)-\sum^{C^l+C^u}_{k=1}\bm{p}^l_i(k)\log\bm{p}^u_j(k).
\end{split}
\end{equation}
In this more efficient way, the inter-class sKLD loss term over a mini-batch can be calculated faster with the matrix multiplication on GPU, promoting our method to be better applied on large-scale datasets (\eg, ImageNet).

\end{document}